\documentclass{article}

\PassOptionsToPackage{numbers, compress}{natbib}

 \usepackage[preprint]{neurips_2025}
 \usepackage{adjustbox} 
\usepackage{array}     

\usepackage[utf8]{inputenc} 
\usepackage[T1]{fontenc}    
\usepackage{hyperref}       
\usepackage{url}            
\usepackage{booktabs}       
\usepackage{amsfonts}       
\usepackage{nicefrac}       
\usepackage{microtype}      
\usepackage{xcolor}         
\usepackage{amsmath}
\usepackage{graphicx}
\usepackage{tabularx}
\usepackage{multirow}

\usepackage{subcaption} 
\usepackage{soul} 

\definecolor{pastelgreen}{HTML}{DFF0D8} 
\definecolor{pastelred}{HTML}{F2DEDE}   

\newcommand{\best}[1]{\sethlcolor{pastelgreen}\hl{#1}}
\newcommand{\second}[1]{\sethlcolor{pastelred}\hl{#1}}

\makeatletter
\newcommand\approxsim{\mathpalette\@approxsim\relax}
\newcommand\@approxsim[2]{%
  \mathrel{%
    \ooalign{%
      $\m@th#1\sim$\cr
      \hidewidth$\m@th#1:$\hidewidth\cr
    }%
  }%
}
\makeatother

\title{Generating healthy counterfactuals with denoising diffusion bridge models}

\author{%
Ana Lawry Aguila$^{1}$ \quad
Peirong Liu$^{1}$\thanks{Currently at the Department of Electrical and Computer Engineering, Johns Hopkins University, Baltimore, USA} \quad
Marina Crespo Aguirre$^{2}$ \quad
Juan Eugenio Iglesias$^{1,3,4}$ \\
\\
$^{1}$Athinoula A. Martinos Center for Biomedical Imaging \\
Massachusetts General Hospital and Harvard Medical School, Boston, USA \\
$^{2}$ETH Zurich, Zurich, Switzerland \\
$^{3}$Computer Science \& Artificial Intelligence Lab \\ Massachusetts Institute of Technology, Boston, USA \\
$^{4}$Hawkes Institute, University College London, London, UK \\
\\
\texttt{acaguila@mgh.harvard.edu}
}

\begin{document}

\maketitle

\begin{abstract}
Generating healthy counterfactuals from pathological images holds significant promise in medical imaging, e.g., in anomaly detection or for application of analysis tools that are designed for healthy scans. These counterfactuals should represent what a patient’s scan would plausibly look like in the absence of pathology, preserving individual anatomical characteristics while modifying only the pathological regions. Denoising diffusion probabilistic models (DDPMs) have become popular methods for generating healthy counterfactuals of pathology data. Typically, this involves training on solely healthy data with the assumption that a partial denoising process will be unable to model disease regions and will instead reconstruct a closely matched healthy counterpart. More recent methods have incorporated synthetic pathological images to better guide the diffusion process. However, it remains challenging to guide the generative process in a way that effectively balances the removal of anomalies with the retention of subject-specific features. 
To solve this problem, we propose a novel application of denoising diffusion bridge models (DDBMs) -- which, unlike DDPMs, condition the diffusion process not only on the initial point (i.e., the healthy image), but also on the \emph{final} point (i.e., a corresponding synthetically generated pathological image).  Treating the pathological image as a structurally informative prior enables us to generate counterfactuals that closely match the patient’s anatomy while selectively removing pathology. The results show that our DDBM outperforms previously proposed diffusion models and fully supervised approaches at segmentation and anomaly detection tasks.

\end{abstract}
\section{Introduction}
Many MRI processing and evaluation tools are designed for healthy images. Large pathological structures, such as brain tumours, can cause these tools to fail~\cite{kofler2024}, making the analysis of disease effects and clinical decision-making more difficult. Generating realistic healthy counterfactuals of disease images offers a means to apply a wide array of existing brain processing algorithms to pathological images and may provide insight into the relationship between healthy and pathological regions. These counterfactuals also facilitate additional applications, such as anomaly detection, where the difference between pathology and pseudo-healthy images can be used to generate anomaly maps for lesion localization and segmentation.

Denoising Diffusion Probabilistic Models (DDPMs)~\cite{Ho2020} have recently gained prominence for generating healthy counterfactuals, particularly in the context of anomaly detection~\cite{Pinaya2022,Bercea2024c,Bercea2023b,Graham2023b,graham2023c,durrer2024}. These models learn to reverse a diffusion process, mapping a complex data distribution to a prior of Gaussian noise. In the context of healthy image generation, the diffusion model is typically trained solely on healthy samples, with the assumption that it will be unable to generate diseased, out-of-distribution samples, and will instead reconstruct healthy tissue. To generate healthy counterfactuals, the denoising process is applied to a partially noised image, rather than sampling from the Gaussian prior, such as to retain some defining characteristics about the original image. However, selecting a noise level that both inpaints anomalies while retaining distinctive features of healthy tissue remains a challenge, with current methods often failing to fully inpaint without compromising anatomical details. 

Incorporating synthetic anomalies into the diffusion model framework to better guide the training process offers a promising alternative to unsupervised methods. Recent diffusion model approaches have used pairs of real healthy images and synthetic pseudo-pathology counterparts to improve the performance over unsupervised approaches~\cite{Baugh2024,lawryaguila2025}. Conditioning the denoising steps on these synthetic images helps better guide the reconstruction process. Whilst these methods have shown improvements over purely unsupervised diffusion model methods, they still struggle to consistently inpaint pathology while retaining individual characteristics. 

When both healthy and diseased data are available, a natural choice would be a method that can map between two complex distributions. Denoising Diffusion Bridge Models (DDBMs) have emerged as a promising alternative for such distribution translation tasks. In DDBMs, the diffusion process is conditioned on both the initial and terminal points, thus learning a direct mapping between the two distributions. This is in contrast to DDPMs, which learn to map between a data distribution and a structurally uninformative Gaussian prior. The DDBM approach is particularly well suited to our use case, where we expect a strong structural correspondence between pathological and pseudo-healthy images. In this work, we introduce the first application of DDBMs to healthy image reconstruction and anomaly detection and demonstrate that they outperform standard diffusion model approaches at segmentation and anomaly detection.


\begin{figure*}[t]
    \centering
    \includegraphics[trim={0 0.5cm 0 0.5cm}, clip, width=\columnwidth]{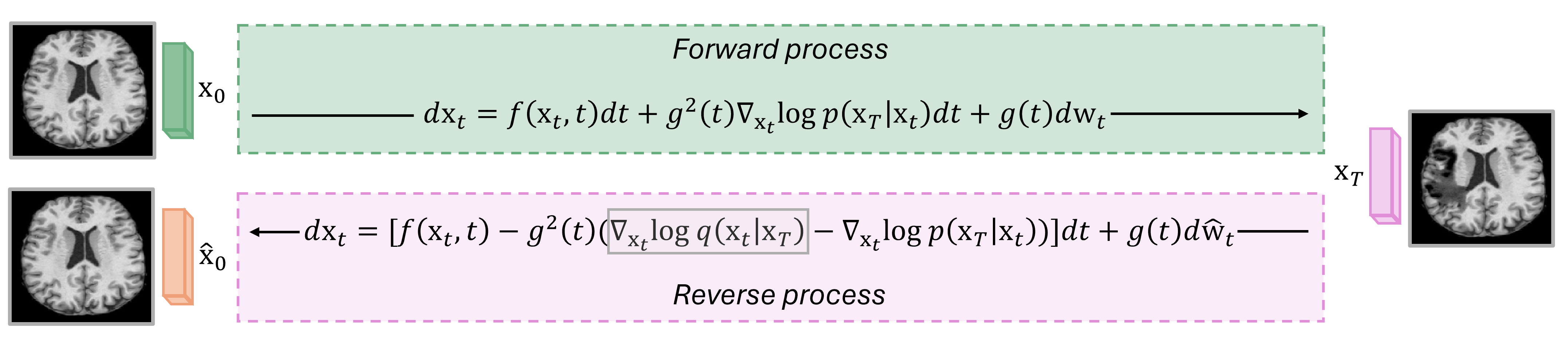}
    \caption{Forward and reverse process of the proposed DDBM framework to map between synthetic pseudo-pathology and real healthy volumes. The denoising bridge score matching objective (Equation \ref{eq:DBSM}) is used to match a parameterized network \( \mathbf{s}_\theta \) to the unknown bridge score function \( \nabla_{\mathbf{x}_t} \log p(\mathbf{x}_t) \).}
    \label{fig:model_framework}
\end{figure*}

\section{Methods}
\subsection{Denoising diffusion probabilistic models}
DDPMs~\cite{Ho2020} define a forward stochastic process that gradually transforms data samples \(\mathbf{x}_0 \sim q(\mathbf{x}_0) \), where \( \mathbf{x} \in \mathbb{R}^d \), into samples from a known prior distribution \( p_T(\mathbf{x}) \), which is generally Gaussian. This transformation is achieved via a time-indexed sequence of variables \( \{ \mathbf{x}_t \}_{t=0}^T \) governed by a linear stochastic differential equation (SDE):
\begin{equation}\label{eq:forward_SDE}
    d\mathbf{x}_t = \mathbf{f}(\mathbf{x}_t,t)dt + g(t)d\mathbf{w}_t,
\end{equation}
where \( \mathbf{f} : \mathbb{R}^d \times [0,T] \rightarrow \mathbb{R}^d \) is the drift function, \( g : [0,T] \rightarrow \mathbb{R} \) is the diffusion coefficient, and \( \mathbf{w}_t \) is a Wiener process. For suitable choices of \( \mathbf{f} \) and \( g \), the transition kernel \( p(\mathbf{x}_t \mid \mathbf{x}_0) \sim \mathcal{N}(\alpha_t \mathbf{x}_0, \sigma_t^2 \mathbf{I})\) is a Gaussian with parameters controlled by \( \alpha_t\) and \( \sigma_t\), time-dependent functions that regulate the signal attenuation and standard deviation in the noising process, respectively. These choices in the forward process ensure that the terminal distribution is approximately Gaussian \( p_T(\mathbf{x}) \approxsim \mathcal{N}(0, \sigma_T^2 \mathbf{I}) \).

To sample from \( q(\mathbf{x}_0) \), we can solve the reverse-time SDE:
\begin{equation} \label{eq:sde}
    d\mathbf{x}_t = \left[\mathbf{f}(\mathbf{x}_t,t) - g^2(t) \nabla_{\mathbf{x}_t}\log p(\mathbf{x}_t)\right]dt + g(t)\, d\hat{\mathbf{w}}_t,
\end{equation}
which shares the same marginals \( \{ p(\mathbf{x}_t) \}_{t=0}^T \) as the forward process and $d\hat{\mathbf{w}}_t$ is the reverse-time Weiner process. The score function \( \nabla_{\mathbf{x}_t} \log p(\mathbf{x}_t) \) can be approximated using a neural network \( \mathbf{s}_\theta \), trained via the denoising score matching objective:
\begin{equation}
    \mathcal{L}(\theta) = \mathbb{E}_{\mathbf{x}_t \sim p(\mathbf{x}_t \mid \mathbf{x}_0),\, \mathbf{x}_0 \sim q(\mathbf{x}_0),\, t \sim \mathcal{U}(0, T)} \left[ \left\| \mathbf{s}_\theta(\mathbf{x}_t, t) - \nabla_{\mathbf{x}_t} \log p(\mathbf{x}_t \mid \mathbf{x}_0) \right\|^2 \right]
\end{equation}
 
which is tractable because the transition kernel has known mean and variance from the forward SDE. 

\subsection{Denoising diffusion bridge models}\label{sec:diffusion_bridges}
Denoising diffusion bridge models (DDBMs)~\cite{Zhou2024} extend the diffusion model framework by considering both initial and terminal states as drawn from a joint data distribution $(\mathbf{x}_0,\mathbf{x}_T) \sim q(\mathbf{x}_0,\mathbf{x}_1)$. In this framework, the endpoint $\mathbf{x}_T = \mathbf{x}_1$ is no longer restricted to Gaussian noise but is chosen as an informative target from the data distribution and reversing the process involves sampling from \(q(\mathbf{x}_t \mid \mathbf{x}_T)\). Note that the distribution $q(\cdot)$ differs from the DDPM marginal distribution $p(\cdot)$, as the endpoint distributions are now given by $q(\mathbf{x}_0, \mathbf{x}_T) = q(\mathbf{x}_0, \mathbf{x}_1)$ rather than $p(\mathbf{x}_0, \mathbf{x}_T) = p(\mathbf{x}_T | \mathbf{x}_0)q(\mathbf{x}_0)$~\cite{Zhou2024}. The DDBM diffusion process is achieved by modifying the forward SDE using Doob's $h$-transform~\cite{Doob1985}:  
\begin{equation}\label{eq:forward_SDE_bridge}
  d\mathbf{x}_t = \mathbf{f}(\mathbf{x}_t,t)\,dt + g^2(t)\nabla_{\mathbf{x}_t} \log p(\mathbf{x}_T  \mid \mathbf{x}_t)\,dt + g(t)\,d\mathbf{w}_t
\end{equation}
where $p(\mathbf{x}_T \mid \mathbf{x}_t)$ is the transition kernel of the underlying diffusion process. For linear SDEs, this forward bridge process gives rise to a tractable Gaussian bridge kernel:
\begin{equation}\label{eq:kernel_ddbm}
\begin{aligned}
q(\mathbf{x}_t \mid \mathbf{x}_0, \mathbf{x}_T) &\sim \mathcal{N}\left(a_t \mathbf{x}_T + b_t \mathbf{x}_0,\, c_t^2 \mathbf{I} \right), \\
a_t &= \frac{\alpha_t\, \mathrm{SNR}_T}{\alpha_T\, \mathrm{SNR}_t}, \quad
b_t = \alpha_t \left(1 - \frac{\mathrm{SNR}_T}{\mathrm{SNR}_t} \right), \quad
c_t^2 = \sigma_t^2 \left(1 - \frac{\mathrm{SNR}_T}{\mathrm{SNR}_t} \right)
\end{aligned}
\end{equation}
with signal-to-noise ratio $\text{SNR}_t=\alpha_t^2/\sigma_t^2$. The reverse SDE is given by:
\begin{equation}\label{eq:ddbm_sde}
    d\mathbf{x}_t = \left[ f(\mathbf{x}_t, t) - g^2(t) \left( \nabla_{\mathbf{x}_t} \log q(\mathbf{x}_t\mid \mathbf{x}_T) - \nabla_{\mathbf{x}_t} \log p(\mathbf{x}_T \mid \mathbf{x}_t) \right) \right] dt + g(t) d\hat{\mathbf{w}}_t,
\end{equation}
which shares the same marginals $\left\{ q(\mathbf{x}_t \mid \mathbf{x}_T) \right\}_{t=0}^T$ as the forward process. The bridge score function $\nabla_{\mathbf{x}_t} \log q(\mathbf{x}_t \mid \mathbf{x}_T)$ can be approximated using the denoising bridge score matching objective:
\begin{equation}
\label{eq:DBSM}
\mathcal{L}_w(\theta) = 
\mathbb{E}_{\mathbf{x}_0, \mathbf{x}_T, \mathbf{x}_t, t}
\left[ w(t) \left\| s_\theta(\mathbf{x}_t, \mathbf{x}_T, t) - \nabla_{\mathbf{x}_t} \log q(\mathbf{x}_t \mid \mathbf{x}_0, \mathbf{x}_T) \right\|_2^2 \right]
\end{equation}
where $w(t)$ is a time-dependent weighting function and $q(\mathbf{x}_t \mid \mathbf{x}_0, \mathbf{x}_T)$ is the tractable bridge kernel from Equation \ref{eq:kernel_ddbm}. In this work, we speed up sampling by using the diffusion bridge implicit model (DBIM) sampling procedure proposed by~\cite{Zheng2025}: 
\begin{equation}
\mathbf{x}_{t_n} = a_{t_n} \mathbf{x}_T + b_{t_n} \hat{\mathbf{x}}_0 + 
\sqrt{c_{t_n}^2 - \rho_n^2}
\hat{\boldsymbol{\epsilon}}
+ \rho_n \boldsymbol{\epsilon}, \quad 
\hat{\boldsymbol{\epsilon}} =  \frac{\mathbf{x}_{t_{n+1}} - a_{t_{n+1}} \mathbf{x}_T - b_{t_{n+1}} \hat{\mathbf{x}}_0}{c_{t_{n+1}}}, \quad
\boldsymbol{\epsilon} \sim \mathcal{N}(0, \mathbf{I})
\end{equation}

where $\rho \in \mathbb{R}^{N-1}$, $\hat{\boldsymbol{\epsilon}}$ denotes the predicted noise and $n$ denotes the discrete timestep. \(\hat{\mathbf{x}}_0=\mathbf{x}_\theta(\mathbf{x}_{t_{n+1}}, t_{n+1}, \mathbf{x}_T)\) is the predicted denoised data at time \( t = 0\) and \(\mathbf{x}_\theta\) is the data predictor network. This can can be related to the score function by \(s_\theta(\mathbf{x}_t, t, \mathbf{x}_1)=-[\mathbf{x}_t - a_t · \mathbf{x}_1 - b_t · \hat{\mathbf{x}}_0] / c_t^2 · \).

\subsection{DDBMs for healthy counterfactual generation}\label{sec:diffusion_bridges_for_healthy}
In this work, we apply a DDBM to map from pathological to healthy images, utilizing synthetic disease images and their corresponding ground-truth healthy counterparts. To enable this, we first need a method for generating synthetic pathology from healthy images. For this, we use a recently proposed fluid-driven anomaly randomization framework (``UNA''~\cite{Liu2025}). This method models pathology as an advection-diffusion process governed by partial differential equations, generating well-defined synthetic anomalies through controllable velocity fields and boundary conditions. The initial anomaly condition is derived from real lesion segmentations, ensuring that the generated anomalies are realistic. The resultant synthetic pathological–healthy image pairs are used to train the DDBM, with the healthy image serving as the start point $\mathbf{x}_0$ and the pathological image as the target endpoint $\mathbf{x}_T$. An overview of our framework is shown in Figure \ref{fig:model_framework}. We implement $s_\theta$ in Equation \ref{eq:DBSM} by extending the UNet architecture of~\cite{Song2021} to 3D and sample in reverse time across 10 steps using the DBIM sampling procedure. Further implementation details are available in the supplemental.

\section{Experimental setup}
We evaluate our DDBM on two downstream tasks. The first goal is to generate healthy counterfactuals that preserve the structure of healthy regions surrounding pathology. In synthetic pathology scenarios, the generated counterfactuals should closely resemble the corresponding ground-truth healthy images. Rather than focusing on pixel-wise similarity, we assess their usefulness for downstream analysis. Specifically, we evaluate whether pseudo-healthy images retain individual-specific characteristics when processed with existing image analysis tools. For the test cohorts of each dataset described in Section \ref{sec:datasets}, we generate healthy counterfactuals using our method as well as the comparison baselines. The SynthSeg~\cite{Billot2023} segmentation algorithm is then run on both the ground-truth healthy images and the generated counterfactuals, and the resulting segmentation maps are compared. 

For the specific application of translating pathology images into healthy counterfactuals, a second goal is to ensure that the generated images are free of pathological tissue. To evaluate this, we apply our method to an anomaly detection task, evaluating its ability to inpaint real lesions from ATLAS -- a dataset of brain MRI scans with strokes. For the unseen ATLAS test cohort, we generate counterfactuals and build anomaly maps by comparing pathology images with their corresponding counterfactuals, following the approach of~\cite{lawryaguila2025}.

\subsection{Datasets and comparison methods}\label{sec:datasets}
We train a DDBM using T1-weighted MRI scans (with image processing steps and cohort sizes described in the supplement) from the following datasets; ADNI~\cite{Weiner2017}, HCP~\cite{Essen2012}, ADHD200~\cite{Brown2012}, AIBL~\cite{Fowler2021}, and OASIS3~\cite{LaMontagne2018}. While ADNI, AIBL, and OASIS3 contain older subjects (with white matter hyperintensities or atrophy) and ADHD200 includes individuals diagnosed with ADHD, we treat all these scans as approximately healthy, as they do not have larger lesions that noticeably change the structure of the brain (e.g., tumors or strokes). We generate pseudo-pathology scans for each sample using the fluid-driven anomaly randomization approach described in Section~\ref{sec:diffusion_bridges_for_healthy}. For the initial anomaly condition, we use manual chronic strokes lesion segmentations from ATLAS~\cite{Liew2017}, splitting the dataset into a training and test cohort.

We compare our DDBM approach to the following healthy image reconstruction methods; a Latent Diffusion Model (LDM)~\cite{Rombach2021} previously implement for healthy counterfactual generation~\cite{lawryaguila2025}, a conditional LDM (cLDM) as implemented by~\cite{lawryaguila2025}, and UNA~\cite{Liu2025}, which is a general-purpose model for diseased-to-healthy image generation. We train all methods using the same training cohort.

\begin{table*}
    \centering
    \caption{Segmentation Dice scores for key brain regions. Hippo = Hippocampus, Amyg = Amygdala, Thal = Thalamus, Caud = Caudate,
    Put = Putamen, Pal = Pallidum, Ctx = Cerebral cortex, WM = Cerebral white matter,
    LatVent = Lateral ventricle, 3rdVent = 3rd ventricle. \best{green} and \second{red} indicate the best and second-best results. The average ranks are computed by assigning each method a rank per brain region and then averaging these ranks across all regions within a dataset.}
    \label{tab:dice_results}
    \footnotesize 
    \setlength{\tabcolsep}{2pt} 
    \renewcommand{\arraystretch}{1.1} 
    \begin{tabular}{p{1.3cm}p{1.8cm}
        *{10}{>{\centering\arraybackslash}p{0.9cm}}}
        \toprule
        Dataset & Method 
        & Hippo & Amyg & Thal & Caud & Put & Pal & Ctx & WM & LatVent & Rank\\
        \midrule
        \multirow{4}{*}{HCP}  
            & UNA & \second{0.9035} & \second{0.8907} & 0.9220 & \second{0.9134} & \second{0.9152} & \best{0.8917} & \best{0.9056} & 0.8646 & 0.8932 & \second{2.11} \\
            & LDM & 0.7730 & 0.8170 & 0.8949 & 0.8360 & 0.8684 & 0.8576 & 0.5871 & 0.6856 & 0.7429 & 4.00 \\
            & cLDM & 0.8931 & 0.8815 & \second{0.9304} & 0.9025 & 0.8854 & 0.8643 & 0.8439 & \second{0.9021} & \second{0.9164} & 2.67 \\
            & DDBM (Ours) & \best{0.9072} & \best{0.8994} & \best{0.9441} & \best{0.9291} & \best{0.9192} & \second{0.8917} & \second{0.8614} & \best{0.9202} & \best{0.9340} & \best{1.11} \\
        \midrule
        \multirow{4}{*}{ADNI}  
            & UNA & \second{0.8877} & \best{0.8822} & \second{0.9231} & \second{0.9001} & \second{0.8966} & \second{0.8714} & \second{0.8375} & \best{0.9113} & 0.9426 & \second{1.89} \\
            & LDM & 0.7407 & 0.7792 & 0.8644 & 0.7569 & 0.8300 & 0.8275 & 0.5452 & 0.6716 & 0.8199 & 4.00 \\
            & cLDM & 0.8761 & 0.8649 & 0.9179 & 0.8676 & 0.8768 & 0.8476 & 0.8131 & \second{0.8883} & \second{0.9504} & 2.78 \\
            & DDBM (Ours) & \best{0.8928} & \second{0.8821} & \best{0.9346} & \best{0.9051} & \best{0.9003} & \best{0.8732} & \best{0.8937} & 0.8517 & \best{0.9606} & \best{1.33} \\
        \midrule
        \multirow{4}{*}{ADHD200}  
            & UNA & \second{0.9006} & \second{0.8833} & 0.9291 & \second{0.9165} & \second{0.9254} & \best{0.9153} & \best{0.8710} & \best{0.9197} & 0.8876 & \second{1.89} \\
            & LDM & 0.7631 & 0.7965 & 0.8911 & 0.8262 & 0.8665 & 0.8562 & 0.5988 & 0.6500 & 0.7178 & 4.00 \\
            & cLDM & 0.8870 & 0.8757 & \second{0.9336} & 0.9059 & 0.8979 & 0.8792 & 0.8368 & 0.8844 & \second{0.9093} & 2.78 \\
            & DDBM (Ours) & \best{0.9096} & \best{0.8938} & \best{0.9454} & \best{0.9314} & \best{0.9305} & \second{0.9055} & \second{0.8642} & \second{0.9128} & \best{0.9307} & \best{1.33} \\
        \midrule
        \multirow{4}{*}{OASIS3}  
            & UNA & \second{0.8894} & \second{0.8803} & \second{0.9271} & \second{0.9039} & \second{0.9031} & \best{0.8797} & \best{0.8959} & 0.8638 & 0.9378 & \second{2.00} \\
            & LDM & 0.7347 & 0.7760 & 0.8625 & 0.7726 & 0.8399 & 0.8329 & 0.5415 & 0.6728 & 0.8129 & 4.00 \\
            & cLDM & 0.8767 & 0.8750 & 0.9238 & 0.8806 & 0.8827 & 0.8565 & 0.8194 & \second{0.8946} & \second{0.9499} & 2.78 \\
            & DDBM (Ours) & \best{0.8980} & \best{0.8909} & \best{0.9371} & \best{0.9130} & \best{0.9071} & \second{0.8775} & \second{0.8456} & \best{0.9165} & \best{0.9614} & \best{1.22} \\
        \midrule
        \multirow{4}{*}{AIBL}  
            & UNA & \second{0.8860} & \second{0.8784} & \second{0.9238} & \second{0.9002} & \second{0.9012} & \best{0.8823} & \second{0.8923} & \second{0.8606} & 0.9389 & \second{2.00} \\
            & LDM & 0.7403 & 0.7758 & 0.8624 & 0.7637 & 0.8357 & 0.8300 & 0.5399 & 0.6704 & 0.8120 & 4.00 \\
            & cLDM & 0.8760 & 0.8748 & 0.9214 & 0.8796 & 0.8830 & 0.8517 & 0.8147 & \best{0.8919} & \second{0.9493} & 2.67 \\
            & DDBM (Ours) & \best{0.8937} & \best{0.8873} & \best{0.9350} & \best{0.9077} & \best{0.9016} & \second{0.8714} & \best{0.8927} & 0.8525 & \best{0.9586} & \best{1.33} \\
        \bottomrule
    \end{tabular}
\end{table*}

\subsection{Segmentation results: counterfactuals vs. ground-truth}\label{sec:segmentation} Table \ref{tab:dice_results} shows the Dice scores between the ground-truth healthy and pseudo-healthy segmentations for aggregate SynthSeg regions. Our method achieves the highest Dice for most regions and the highest rank across all datasets. This suggests that our method is best able to generate healthy counterfactuals which retain the defining characteristics of healthy tissue. 
\begin{table*}
    \centering
    \caption{Anomaly detection metrics were computed from anomaly maps and manual annotations, including pixel-wise AUC ($\text{AUC}_{\text{pix}}$), average precision ($\text{AP}_{\text{pix}}$), maximum Dice per sample, and the false positive rate (FPR) based on Dice index threshold.}
    \label{tab:pathology_results}
    \begin{tabularx}{1.\textwidth}{
       X
        >{\centering\arraybackslash}p{1.8cm}
        >{\centering\arraybackslash}p{1.8cm}
        >{\centering\arraybackslash}p{1.8cm}
        >{\centering\arraybackslash}p{1.8cm}
        >{\centering\arraybackslash}p{1.8cm}
    }
        \toprule
        Method & Dice $(\uparrow)$  & $\text{AP}_{\text{pix}}$ $(\uparrow)$ & $\text{AUC}_{\text{pix}}$ $(\uparrow)$ & FPR $(\downarrow)$ & Rank $(\downarrow)$\\
        \midrule
        UNA & \second{0.2614} & \second{0.2210} & \best{0.9539} & \second{0.0100} & \second{2.25} \\
        LDM &  0.2418 & 0.1889 & 0.9303 & 0.0124 & 3.50 \\
        cLDM &  0.2588 & 0.2091 & 0.9236 &  0.0146 & 3.50 \\
        DDBM (Ours) & \best{0.4774} & \best{0.4482} & \second{0.9538} & \best{0.0026} & \best{1.25} \\
        \bottomrule
    \end{tabularx}
\end{table*}

\begin{figure*}
    \centering
    \begin{subfigure}{0.45\linewidth}
        \centering
         \includegraphics[trim={0 0.5cm 0 0.5cm}, clip, width=\linewidth]{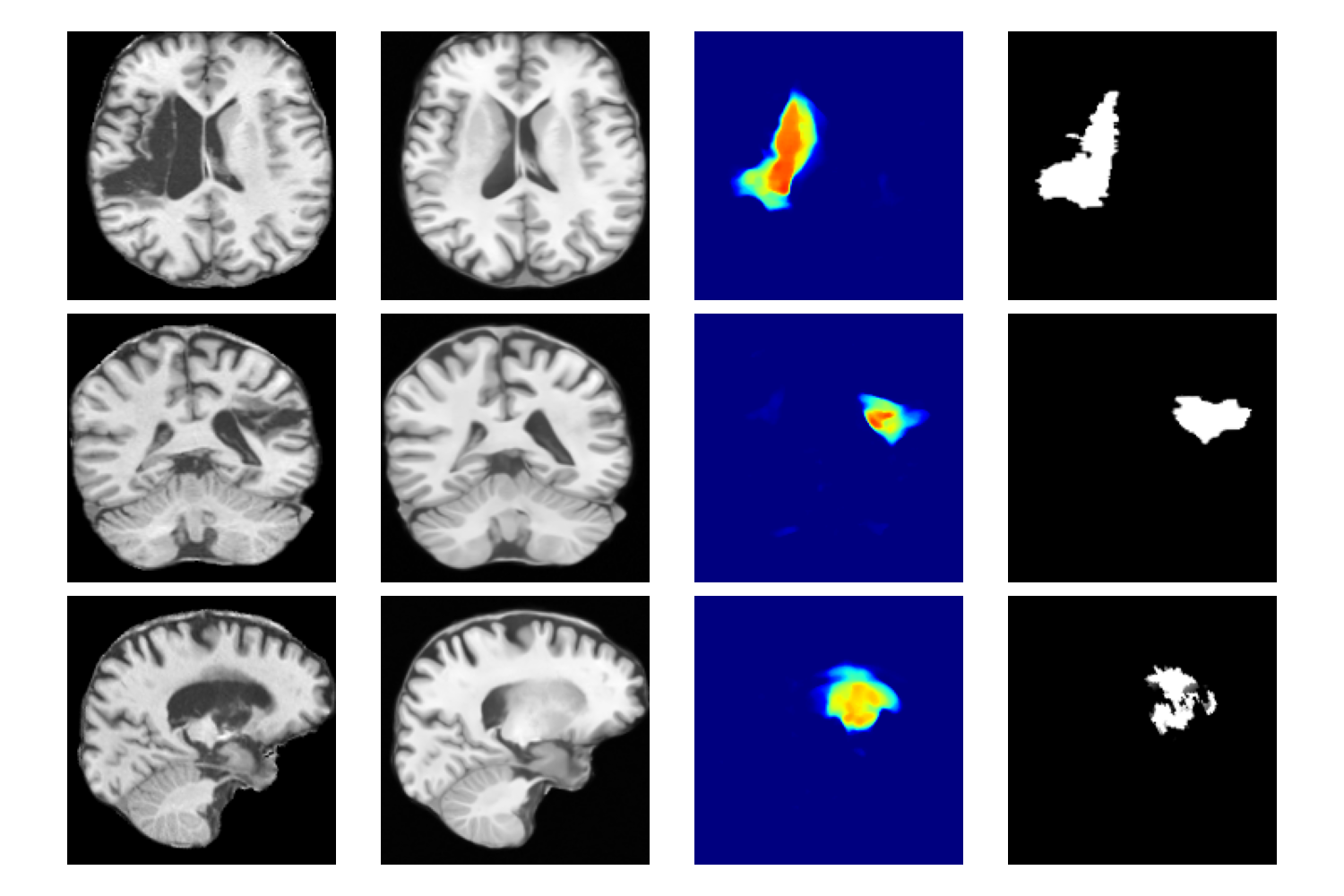}
        \caption{Example 1}
    \end{subfigure}
    \begin{subfigure}{0.45\linewidth}
        \centering
         \includegraphics[trim={0 0.5cm 0 0.5cm}, clip, width=\linewidth]{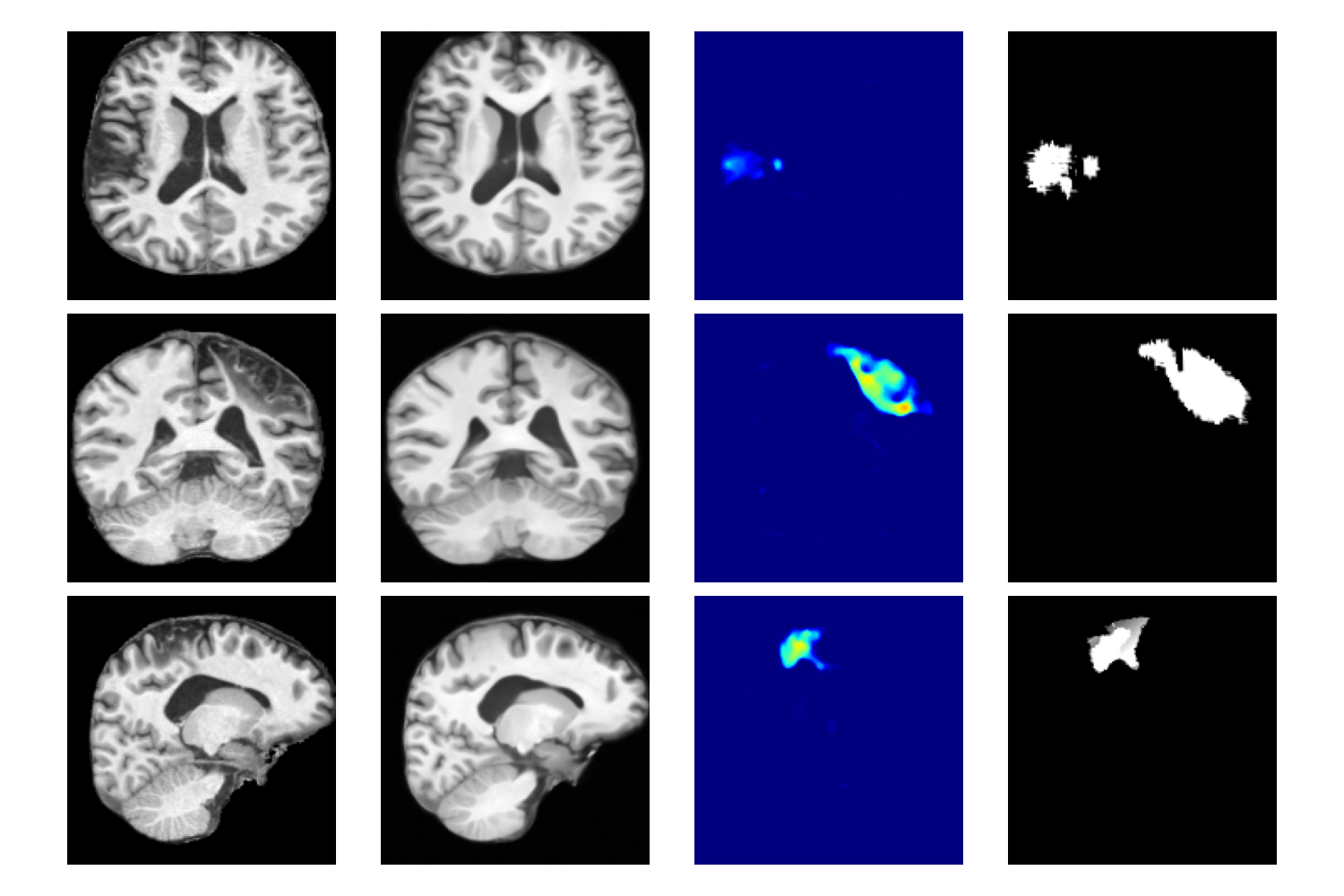}
        \caption{Example 2}
    \end{subfigure} 
    \caption{Qualitative examples of anomaly detection  on test subjects from the ATLAS dataset. Each subfigure shows, from left to right: the original pathological scan, the healthy counterfactual generated by the DDBM, the corresponding anomaly maps, and the ground-truth manual segmentation.}
    \label{fig:qualitative_results}
\end{figure*}

\subsection{Anomaly detection in real pathology}\label{sec:anomaly_detection}
Table \ref{tab:pathology_results} presents the quantitative pathology detection results. Our model outperforms both the unsupervised DDPM and self-supervised cDDPM methods across all metrics, and surpasses the supervised UNA~\cite{Liu2025} approach on all but one metric, achieving the highest overall rank. Qualitative results in Figure \ref{fig:qualitative_results} show that our method generates high-contrast anomaly maps in abnormal regions and low contrast in healthy regions, effectively localizing anomalies while preserving information from the surrounding healthy tissue.

\section{Conclusion}

This work presents a novel application of DDBMs for healthy counterfactual generation, a key task in medical imaging. Our approach outperforms both standard diffusion models and fully supervised baselines on brain tissue segmentation and anomaly detection. The results show that DDBMs effectively capture structural relationships between diseased and healthy scans, producing counterfactuals that inpaint diseased regions while preserving individual anatomical characteristics. Future work will explore whether incorporating anatomical guidance further enhances realism. Overall, DDBMs offer a powerful new avenue for precise medical image counterfactuals, bridging the gap between pathology and healthy anatomy.

\section{Acknowledgements}
This work was primarily funded by NIH grant 1RF1AG080371. Additional support was provided by NIH grants 1RF1MH123195, 1R01AG070988, 1R01EB031114, 1UM1MH130981, and 1R21NS138995. 
\newpage
\bibliographystyle{splncs04}
\bibliography{main}

\newpage
\appendix

\section{Technical Appendices and Supplementary Material}

\subsection{DDBM implementation details}
To implement the Denoising Diffusion Bridge Model (DDBM), we use the codebase provided by~\cite{Zheng2025}. We train the DDBM using the VP noise schedule and a network with the UNet structure proposed by~\cite{Song2021}. Our UNet consists of three levels (128, 256, 256 channels), residual blocks at each level, and attention at the first layer of the decoder. We use the RAdam optimizer with a learning rate of 0.0001 and no weight decay. We use a batch size of 1 and train on a single 80G NVIDIA A100 GPU using gradient checkpointing and mixed precision to reduce computational costs~\cite{Liu2021}.

For inference, we use the DDIM sampling scheme~\cite{Zheng2025} with 10 steps. For baseline methods, we use the code from the original implementation. 

\subsection{Data pre-processing}

To train our method, we use healthy-pseudo-pathology pairs of T1w MRI volumes from the following datasets: ADNI~\cite{Weiner2017} ($N_{\text{train}}$=270, $N_{\text{val}}$=15,$N_{\text{test}}$=31), HCP~\cite{Essen2012} ($N_{\text{train}}$=701, $N_{\text{val}}$=38,$N_{\text{test}}$=82), ADHD200~\cite{Brown2012} ($N_{\text{train}}$=706, $N_{\text{val}}$=32,$N_{\text{test}}$=82), AIBL~\cite{Fowler2021} ($N_{\text{train}}$=545, $N_{\text{val}}$=32,$N_{\text{test}}$=64), and OASIS3~\cite{LaMontagne2018} ($N_{\text{train}}$=1057, $N_{\text{val}}$=53,$N_{\text{test}}$=123). The test cohorts are used in the tissue segmentation analysis in Section \ref{sec:segmentation}. Each healthy image is skull-stripped and bias-field corrected with FreeSurfer~\cite{Fischl2012} and min-max normalized to [0,1], All volumes are affinely registered to MNI152 space using EasyReg~\cite{Iglesias2023b,Hoffmann2022} and transformed and cropped to $160^3$ voxels. For each test sample, we generate 10 synthetic pathology images. To simulate pathology and generate pseudo-pathology scans, we use manual chronic strokes lesion segmentations from the ATLAS dataset~\cite{Liew2017} using $N_{\text{train}}$=590 for training and $N_{\text{test}}$=56 for the anomaly detection analysis in Section \ref{sec:anomaly_detection}.

\end{document}